\documentclass{article}

\usepackage[preprint,nonatbib]{neurips_2023}

\usepackage[numbers]{natbib}
\usepackage{colortbl}
\usepackage{microtype}
\usepackage{hyperref}
\usepackage{url}
\usepackage{booktabs}
\definecolor{darkblue}{rgb}{0, 0, 0.5}
\hypersetup{colorlinks=true, citecolor=darkblue, linkcolor=darkblue, urlcolor=darkblue}
\usepackage{graphicx}
\usepackage{times}
\usepackage[utf8]{inputenc} 
\usepackage[T1]{fontenc}    
\usepackage{amsfonts}       
\usepackage{nicefrac}       
\usepackage{microtype}      
\usepackage{soul}
\usepackage{multirow}
\usepackage{xcolor}
\usepackage[many]{tcolorbox}
\usepackage[export]{adjustbox}

\usepackage{caption}
\usepackage{subcaption}

\title{Detecting and Mitigating Hateful Content in Multimodal Memes with Vision-Language Models}

\author{%
  Minh-Hao Van \\
  Department of EECS\\
  University of Arkanasas\\
  Fayetteville, AR, USA \\
  \texttt{haovan@uark.edu} \\
  \And
  Xintao Wu \\
  Department of EECS\\
  University of Arkanasas\\
  Fayetteville, AR, USA \\
  \texttt{xintaowu@uark.edu} \\
}

\begin{document}

\maketitle

\begin{abstract}
The rapid evolution of social media has provided enhanced communication channels for individuals to create online content, enabling them to express their thoughts and opinions. Multimodal memes, often utilized for playful or humorous expressions with visual and textual elements, are sometimes misused to disseminate hate speech against individuals or groups. While the detection of hateful memes is well-researched, developing effective methods to transform hateful content in memes remains a significant challenge. Leveraging the powerful generation and reasoning capabilities of Vision-Language Models (VLMs), we address the tasks of detecting and mitigating hateful content. This paper presents two key contributions: first, a definition-guided prompting technique for detecting hateful memes, and second, a unified framework for mitigating hateful content in memes, named \texttt{UnHateMeme}, which works by replacing hateful textual and/or visual components. With our definition-guided prompts, VLMs achieve impressive performance on hateful memes detection task. Furthermore, our \texttt{UnHateMeme} framework, integrated with VLMs, demonstrates a strong capability to convert hateful memes into non-hateful forms that meet human-level criteria for hate speech and maintain multimodal coherence between image and text. Through empirical experiments, we show the effectiveness of state-of-the-art pretrained VLMs such as LLaVA, Gemini and GPT-4o on the proposed tasks, providing a comprehensive analysis of their respective strengths and limitations for these tasks. This paper aims to shed light on important applications of VLMs for ensuring safe and respectful online environments.\\
\textcolor{red}{Warning: This paper contains examples of hateful content.}
\end{abstract}

\section{Introduction}
\label{sec:intro}
Nowadays, social media platforms serve as extensive repositories for a wide range of content, including text, images, audio, and video. However, this vast array of information presents a dual challenge: it may not always be safe for viewers, potentially containing explicit or implicit harm to viewers. Multimodal memes, often utilized for playful or humorous expression with visual and textual elements, are sometimes misappropriated to disseminate hate speech against individuals or groups based on their characteristics. Consequently, identifying hateful memes has become a critical issue that social media platforms aim to address to protect users \citep{kiela2020hateful}. Detecting hateful memes, however, is a non-trivial task since multimodal memes combine visual and textual components to convey messages. The task is more challenging when the hate can be either unimodal or multimodal. While unimodal hate can be identified by assessing the hatefulness in either the image or the text modality, multimodal hate is more difficult to detect because the image and text are non-hateful individually but together convey a hateful message. Further, the hateful content is not always explicitly expressed but is implicitly hidden within a historical story behind it. Therefore, intelligent systems are urgently needed to process the vast digital content uploaded to social platforms. Many deep learning techniques have been proposed for this detection task \citep{he2016deep,ren2015faster,devlin2018bert,kiela2019supervised,lu2019vilbert,li2019visualbert}. While significant research has been dedicated to detecting hateful memes, there is a notable gap in methodologies for converting unintentionally hateful memes into more appropriate versions. Addressing this gap represents the next important step in hate speech research, highlighting the potential to assist online users in creating safe and lighthearted memes to be shared on social media.

Large language models (LLMs) have demonstrated remarkable performance across various natural language processing tasks in recent years. With a strong ability to learn human language from vast corpora of text, LLMs can encode enormous amounts of knowledge as embedding vectors, which is a backbone for transforming the knowledge to different forms of representations with the help of adapters \citep{zhang2023llama} or decoder modules \citep{wu2023next}. This flexibility allows AI practitioners to combine text with other sources of information to build multimodally large models. When integrated with vision capabilities, LLMs enable the exploration of emergent abilities in multimodal contexts. Vision-language models (VLMs), such as CLIP, LLaVA, Flamingo, GPT-4, and Gemini, have demonstrated impressive performance across various vision-linguistic tasks. VLMs and their variations can not only perform well on general tasks but also be applied in other fields such as medical data \citep{li2023llava,singhal2022large,van2024large}. While VLMs have gained traction in different fields, socially responsible problems remain under-explored.

In this paper, we propose VLM-based approaches for detecting and mitigating hateful memes, making two key contributions: (1) a definition-guided prompting technique for detecting hateful memes and (2) a unified framework for mitigating hateful content in memes. To the best of our knowledge, our work is among the first few works to leverage prompting techniques with VLMs for the detection of hateful memes and the first to explore the hateful memes mitigation -- a critical step towards ensuring safe and respectful interactions on online social media environments in the era of LLMs and VLMs. For example, users may share memes intended to be humorous or satirical, but there remains a major concern whether such content, either explicitly or implicitly, conveys harmful messages to viewers. More importantly, if memes are determined to be hateful, an open and intriguing question arises: \textit{what actions can be taken to assist social media users in combating toxic content and promoting positive messages?} As a result, we focus on exploring a novel strategy for modifying hateful memes such that the mitigated versions convey meaningful messages without being offensive, addressing a challenging societal issue that requires interdisciplinary collaboration between AI and social science researchers.

To effectively detect hateful memes, we propose a definition-guided prompt that incorporates a detailed definition of hatefulness and clear classification criteria to enhance the reasoning capabilities of VLMs. With our robust detection prompts, VLMs can outperform multiple hateful memes detection baselines without the need for additional fine-tuning steps or OCR text. Notably, our results reveal that VLMs yield substantial improvements in AUROC scores with our proposed prompts, compared to prior findings with simpler prompts discussed in \citep{alayrac2022flamingo,awadalla2023openflamingo}.

To tackle the challenge of mitigating hateful content, we propose a mitigation framework, named \texttt{UnHateMeme}, for converting hateful memes to non-hateful ones by leveraging VLMs with task-specific prompts. Our framework decomposes the task of mitigating hateful memes into multiple sequential steps, including tasks such as analyzing the type of hate, identifying the source of hate, and generating substitute image or text. We employ GPT-4o as the central reasoning component in our mitigation framework due to its demonstrated superior performance and exceptional generative capabilities, as evidenced by our comprehensive experiments on hateful meme detection. Nonetheless, the framework is modular and can be easily adapted to integrate other state-of-the-art VLMs. To evaluate the hateful memes mitigation task, we first use our unified framework to transform hateful memes into non-hateful versions and create a corresponding dataset of mitigated memes. Human evaluation is then adopted to assess the quality of the mitigated dataset based on two metrics: \textit{non-hatefulness} and \textit{image-text coherence}. With exceptional generative capabilities, we incorporate VLMs like GPT-4o into our framework as the primary reasoning component for the aforementioned tasks. Our experimental results indicate that the \texttt{UnHateMeme} framework effectively transforms hateful memes into non-hateful versions by removing offensive visual and/or textual elements and suggesting new alternatives while ensuring multimodal image-text coherence in the new memes. This marks a substantial advancement over previous deep learning models, which lacked the capacity for such content generation. 
To facilitate research reproducibility, an anonymous folder containing our source code and data is available at \textcolor{blue}{\url{https://tinyurl.com/unhatememe}}.

\section{Related Works}
\label{sec:rel}

\subsection{Hate Speech Detection}
There has been a concerted effort from the AI research community to address hate speech problems, including hate speech detection datasets \citep{mathew2021hatexplain,mollas2020ethos} or hateful memes detection challenge \citep{kiela2020hateful}. To deal with the hate speech detection task, \citep{DBLP:conf/pakdd/XuYWNZZW22} proposed a framework to detect harmful text by evaluating coded words, which silently represent hateful meanings on social media, by applying a two-layer network on contextual embeddings from ELMo \citep{sarzynska2021detecting}. BERT-MRP \citep{kim-etal-2022-hate} proposed a two-stage BERT-based model, including masked rationale prediction (MRP) and hate speech detection. \citep{zhang2022opt} also evaluated Open Pretrain Transformer (OPT) on the hate speech detection task via zero-, one-, or few-shot cases. Besides hateful text, hatefulness can exist in multimodal form. \citep{kiela2020hateful} introduced a challenge about detecting hate speech in multimodal memes. With enormous baselines on three categories: unimodal models from computer vision or natural language processing tasks \citep{he2016deep,ren2015faster,devlin2018bert}, multimodal models by combining smaller unimodally pretrained models \citep{kiela2019supervised,lu2019vilbert,li2019visualbert}, or multimodal models \citep{lu2019vilbert,li2019visualbert}, the challenge still calls for more contribution from communities. Many state-of-the-art methods for detecting hateful memes have been proposed during the challenge. \citep{zhu2020enhance,muennighoff2020vilio,velioglu2020detecting,lippe2020multimodal,sandulescu2020detecting} proposed an ensemble approach of one or multiple vision and language models to get the final prediction.  

Recent efforts have been directed toward hate speech moderation utilizing Large Language Models \cite{hee2024recent}. Prompting methods have been proposed to classify hateful memes with pretrained language models \cite{cao2023prompting,lin2024towards}. Ji et al. \cite{ji2024capalign} proposed incorporating visual captioning and attribute extraction to enhance the zero-shot detection performance of LLMs. In the few-shot learning setting, Cao et al. \cite{cao2024modularized} proposed a modularized network that effectively detects hateful memes by fine-tuning LLMs through parameter-efficient techniques. Furthermore, generative models have been employed to create memes with varying objectives. For example, MemeCraft \cite{wang2024memecraft} was designed to generate memes promoting climate action and gender equality, while UnsafeDiffusion \cite{qu2023unsafe} focused on generating unsafe images and hateful memes.

\subsection{Vision-Language Models}
In addition to traditional deep learning models specifically trained for classification tasks, Vision-Language Models have been catching much attention over many interesting applications. CLIP \citep{radford2021learning} proposed a new network to connect image and text data. The method guides the model in learning visual concepts in classification tasks using supervision from natural language. OpenAI GPT-4 \citep{OpenAI2023GPT4TR}, Google Gemini \citep{team2023gemini}, Flamingo \citep{alayrac2022flamingo}, OpenFlamingo \citep{awadalla2023openflamingo}, and LLaVA \citep{liu2023visual} demonstrated their effectiveness on various vision-language tasks such as question-answering or human-like chatbot. Along with groundbreaking large foundation models, many research works have explored the extension of large models to different domains. 
LLaVA-Med \citep{li2023llava} and Med-PaLM \citep{singhal2022large} can analyze the medical images and answer related questions like a human expert.

\section{Hateful Memes Detection}
\label{sec:meme_detection}

\subsection{Problem Definition} 
\label{sec:pd_meme_detection}
A dataset $\mathcal{D} =(\mathcal{X},\mathcal{Y})$ contains multiple pairs of data sample $(x_i,y_i)$'s, where $x_i\in \mathcal{X}$ is the input meme and $y_i\in \mathcal{Y}$ are label of hateful or non-hateful, respectively. For memes dataset, each input image $x_i$ contains the visual part $v_i$ and the textual part $t_i$, e.g., $\mathcal{X}=(\mathcal{V},\mathcal{T})$. We define the task of hateful memes detection as a classification problem $f: \mathcal{V}\times\mathcal{T} \rightarrow \mathcal{Y}$.

\noindent\textbf{OCR for meme text.} In general, a discrete visual element $v_i$ or textual element $t_i$ is not always available. With online memes, the meme text is often integrated into the image itself, resulting in the final meme $x_i$. Previous research on hateful meme detection has utilized Optical Character Recognition (OCR) methods to extract the meme text, thereby aiding models in better analyzing the hatefulness within the textual modality. While recent state-of-the-art VLMs have shown excellent capabilities in extracting texts from images without the need for additional OCR modules, we also evaluate the setting where OCR-extracted text is available. This approach allows for a comprehensive comparison with previous studies and an assessment of how VLMs perform both with and without the use of OCR.

\noindent\textbf{Challenges with multimodal memes.} Although $f$ is a binary classification task (hateful or non-hateful), accurately detecting hateful memes is non-trivial due to the presence of both unimodal and multimodal hate.  Hateful memes with unimodal hate can be identified by assessing the hate in either image or text individually. However, hateful memes with multimodal hate pose a more complex challenge, as the meme image $v_i$ and text $t_i$ are non-hateful on their own but together convey a hateful message. This necessitates an effective approach for interpreting the message conveyed by both modalities. To address these challenges, we introduce definition-guided prompts to guide VLMs in detecting hateful memes.

\subsection{Definition-guided Prompt} 
\begin{figure}[t]
    \centering
    \begin{tcolorbox}[sidebyside,lefthand width=3.0cm,lower separated = false, left=2pt, right=2pt, top=2pt, bottom=2pt, sidebyside gap=4pt,before skip=0pt]
        \includegraphics[width=\textwidth]{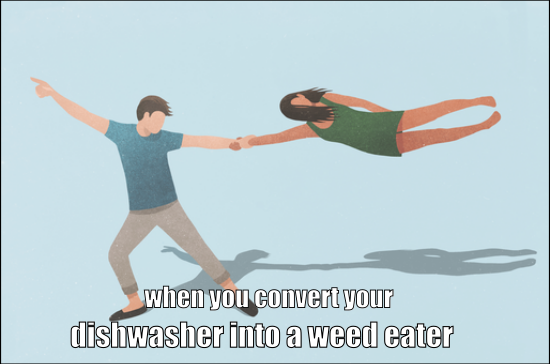}
        \tcblower
        \noindent\textbf{User:} Is this meme hateful? Answer:
        \\
        \textbf{Assistant:} No, this meme isn't hateful. It's a humorous piece depicting a couple dancing, with a playful caption about converting a dishwasher into a weed eater. It seems intended to be lighthearted and silly.
    \end{tcolorbox}
\caption[GPT-4o fails]{An example of GPT-4o failing to detect hateful meme.}
\label{fig:simple_prompt}
\end{figure}

While one can prompt the VLM to enlarge its capabilities on various tasks, it is essential to note that the model's performance dramatically depends on the quality of prompts, including clear instructions for the specific task. Detecting hateful memes presents a unique challenge in crafting an effective prompt. This challenge involves deciphering slang or historical references within visual and textual elements to identify a hateful message. Another challenging problem is how to guide the model to effectively reason about the source of hate in the memes. In Figure \ref{fig:simple_prompt}, we illustrate an example where GPT-4o demonstrates a strong ability to describe the meme, yet it fails to detect hateful meme using a simple prompt as used in \cite{awadalla2023openflamingo}. 

A comprehensive definition of hate is, therefore, necessary. We adapt the definition of hate \citep{kiela2020hateful} as part of the instructions in our prompt. Subsequently, we provide the VLM with guidelines on how to classify memes as hateful or non-hateful by illustrating the strategy for analyzing both the visual and textual components and how to combine them to achieve the desired outcome. In Figure \ref{fig:def_guided_prompt}, we introduce our task-specific zero-shot prompt, referred to as definition-guided prompt, for detecting hateful memes using VLMs.

\subsection{Definition-guided Few-shot Prompt}
Our definition-guided prompt can be adapted to a few-shot setting by incorporating multiple relevant in-context examples to improve detection. The performance of in-context learning heavily relies on the quality of these examples, as those closely related to each test item are more effective representatives for inference. For each test meme, we use Retrieval-based In-Context Example Selection (RICES) \cite{yang2022empirical} to find similar demonstration memes from a candidate set. Since cloud VLMs like Gemini or GPT-4o don't provide visual embeddings, we use the CLIP model's visual encoder to obtain meme embeddings. We then calculate and rank cosine similarities between test and candidate memes to select the best demonstration set, ensuring a balanced number of examples per class. Figure \ref{fig:def-guided-fshot-prompt} shows a prompt example with four demonstrations, with the full prompt detailed in Appendix \ref{sec:apdix_prompts}.

\begin{figure}[t]
    \centering
    \begin{subfigure}[t]{\columnwidth}
        \begin{tcolorbox}[colback=green!15,colframe=green!50!blue!100,left=2pt, right=2pt, top=2pt, bottom=2pt]
        \textbf{User:}
        \noindent Hatefulness definition: Hate speech is defined as a direct or indirect attack on people based on characteristics, including ethnicity, race, nationality, immigration status, religion, caste, sex, gender identity, sexual orientation, disability, disease, mocking hate crime or evoking pain from historical figures. The definition of attack includes violent or dehumanizing speech, statements of inferiority, and calls for exclusion or segregation.
        \\\\
        Classification criteria: A meme is considered hateful if either the image or text is hateful on its own. The meme is also hateful if both the image and text are not hateful individually, but together convey a hateful message. Conversely, if neither the image nor the text is hateful and they do not together create a hateful meaning, the meme is not considered hateful.
        \\\\
        \texttt{<upload-test-meme>}\\
        You consider visual and textual meanings to classify the meme as hateful or non-hateful based on the hatefulness definition and classification criteria.
        After giving a short explanation, provide the answer with either hateful or non-hateful in the following format:\\
        Classification:\\
        Probability of the meme being hateful (from 0 to 1):
        \end{tcolorbox}
        \caption[definition-guided zero-shot prompt]{Definition-guided zero-shot prompt.}
        \label{fig:def_guided_prompt}
    \end{subfigure}
    \begin{subfigure}[t]{\columnwidth}
        \begin{tcolorbox}[colback=green!15,colframe=green!50!blue!100,left=2pt, right=2pt, top=2pt, bottom=2pt] 
        \textbf{User:}
        \texttt{<similar-to-definition-guided-prompt>}
        \\
        Here are some demonstrations on classifying memes:
        \\
        \texttt{<upload-demonstration-meme-1>}\\
        Classification: \texttt{<label-1>}
        \\
        \texttt{<upload-demonstration-meme-2>}\\
        Classification: \texttt{<label-2>}
        \\
        \texttt{<upload-demonstration-meme-3>}\\
        Classification: \texttt{<label-3>}
        \\
        \texttt{<upload-demonstration-meme-4>}\\
        Classification: \texttt{<label-4>}
        \\
        \texttt{<similar-to-definition-guided-prompt>}
    \end{tcolorbox}
    \caption{Definition-guided few-shot prompt.}
    \label{fig:def-guided-fshot-prompt}
    \end{subfigure}    
    \caption{Our prompts for hateful memes detection.}
\end{figure}

\section{Hateful Memes Mitigation}
\label{sec:meme_mitigation}
\begin{figure}[t]
    \centering
    \includegraphics[width=.95\textwidth]{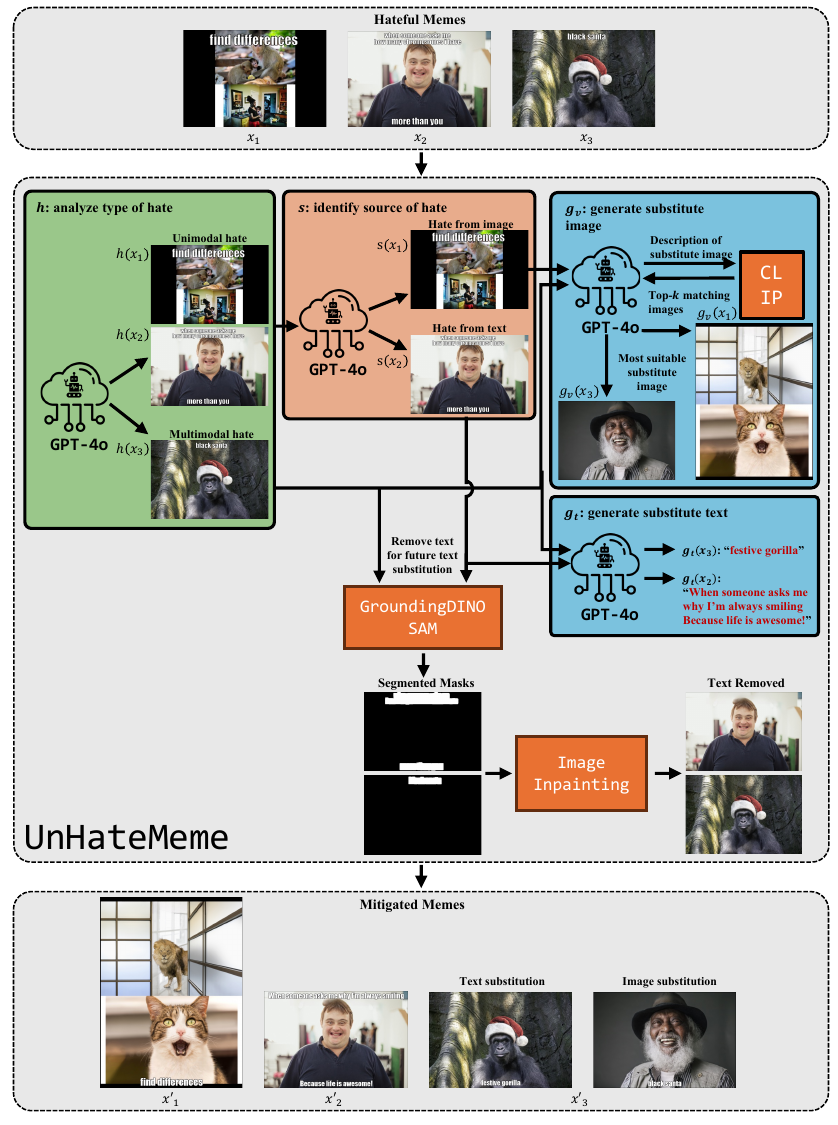}
    \caption{Overview of \texttt{UnHateMeme} framework for mitigating hateful content in memes.}
    \label{fig:mitigation_illustrative}
\end{figure}

\subsection{Problem Definition}
\label{sec:pd_meme_mitigation}
Let $\mathcal{D}_H \subset \mathcal{D}$ be a subset including hateful memes in $\mathcal{D}$. For each $(x_i,y_i)\in \mathcal{D}_H$, $x_i=(v_i,t_i)$ and $y_i=1$ (hateful). We define the task of mitigating hateful content in memes as the problem of finding suitable substitutions, represented by the function $g: \mathcal{V}\times \mathcal{T} \rightarrow \mathcal{V}^\prime \times \mathcal{T}^\prime$. Given that a meme typically consists of both visual and textual components, the function $g$ is designed to address two distinct tasks: $g_t$, responsible for finding a substitute text $t_i^\prime=g_t(v_i,t_i)$, and $g_v$, responsible for finding a substitute image $v_i^\prime=g_v(v_i,t_i)$. The objective is to convert the hateful meme $x_i$ into a modified version $x_i^\prime=(v_i^\prime,t_i^\prime)$, which will be classified as non-hateful, i.e, $y_i^\prime=0$. 

The content generation capabilities of VLMs can be leveraged to address these tasks. Specifically, $g_t$ can be approached as a text generation task by directly prompting GPT-4o to generate a substitute text. Designing the task $g_v$ is more complicated as it requires multiple steps to obtain the substitute image. In particular, $g_v$ involves prompting the VLM to create a description of the potential substitute image, followed by matching this description against a collection of non-hateful images to identify the most suitable substitution.

\noindent\textbf{Mitigation criteria.} The objective of the mitigation task is to modify hateful memes to make them suitable for sharing on social networks. These memes aim to spread playful and lighthearted content without being offensive to other users. Additionally, the framework must ensure coherence between image and text in the newly created meme. For example, a na\"ive method might simply replace the original text in Figure \ref{fig:simple_prompt} with ``\textit{It's a rainy day!}'' -- a substitution that fails to convey any meaningful message when paired with the image. Thus, we anticipate the proposed framework to satisfy two criteria for the mitigated memes: \textit{non-hatefulness} and \textit{image-text coherence}.

\noindent\textbf{Challenges with mitigating hateful content.} Developing an effective generation method $g$ alone is insufficient for a practical application due to the challenge of determining whether the image or text modality requires correction. In other words, users must manually specify either $g_v$ (for image) or $g_t$ (for text), or both, to perform the generation -- a process that ideally should be automated by the mitigation framework. Therefore, we establish that an effective framework for mitigating hateful memes should be unified and capable of performing the task in an end-to-end manner. 

As discussed in Section \ref{sec:pd_meme_detection}, the hate in memes can be either unimodal or multimodal, depending on how the hateful content is embedded in the visual and textual components. The decision on whether to address the image or text modality hinges on the type of hate depicted in the memes. For a meme with unimodal hate, which often explicitly contains hateful image or text, the corresponding hateful components can be corrected to mitigate the hate. Additionally, there are memes in which both the image and text independently express hate. For a meme with multimodal hate, altering either modality removes the hateful message, as it only emerges from the specific combination of image and text. Thus, distinguishing between unimodal and multimodal hate is a crucial step in the framework. Once the type of hate in memes is identified, the next step for unimodal hate cases is to determine which modality (image or text) is responsible for the hateful content. These challenges lead to two additional tasks beyond the main generation task: analyzing the type of hate, denoted as $h: \mathcal{V}\times\mathcal{T} \rightarrow \{\mathrm{unimodal-hate}, \mathrm{multimodal-hate}\}$, and identifying the source of hate, denoted as $s: \mathcal{V}\times\mathcal{T} \rightarrow \{\mathrm{image}, \mathrm{text}, \mathrm{both}\}$.

\subsection{Unified Framework for Mitigating Hateful Content in Memes} 
The objective of mitigating hateful memes involves transforming hateful content into non-hateful versions while preserving the multimodal coherence between visual and textual elements. As aforementioned, a viable framework for mitigating hateful content in memes should automatically handle three tasks: analyzing the type of hate, identifying the source of hate, and generating substitute image or text. To this end, we propose a unified and end-to-end framework called \texttt{UnHateMeme}. This framework processes hateful memes and generates non-hateful counterparts with minimal alterations, ensuring that the content remains as faithful to the original as possible. We leverage the advanced generative capabilities of VLMs, such as GPT-4o, and propose task-specific prompts to guide the model toward achieving the desired outcome. Figure \ref{fig:mitigation_illustrative} offers an illustrative overview of our \texttt{UnHateMeme} framework.

\begin{figure}[ht]
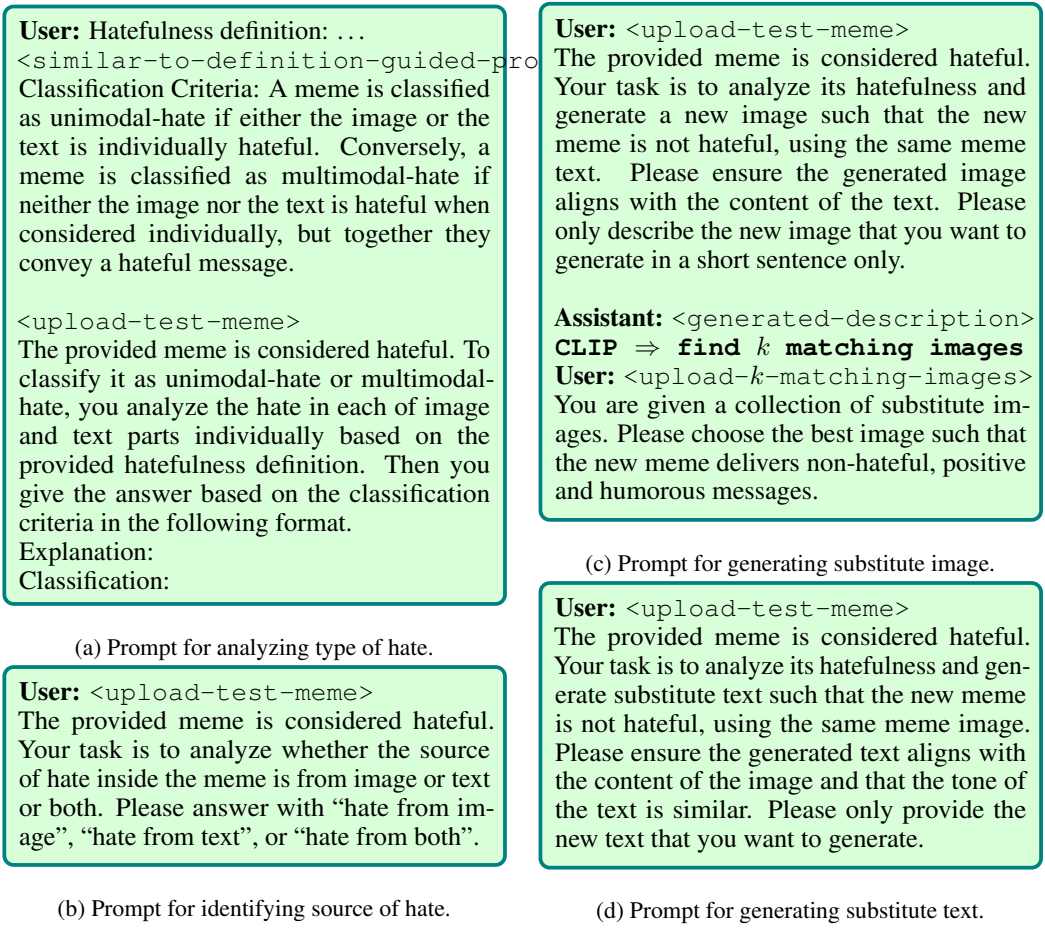

    \centering
    \begin{tabular}{cc}
    \parbox{.48\columnwidth}{
    \begin{subfigure}[t]{.48\columnwidth}
        \begin{tcolorbox}[colback=green!15, colframe=green!50!blue!100,left=2pt, right=2pt, top=2pt, bottom=2pt] 
        \textbf{User:} 
        Hatefulness definition: \dots
        \\ \texttt{<similar-to-definition-guided-prompt>}
        \\
        Classification Criteria: A meme is classified as unimodal-hate if either the image or the text is individually hateful. Conversely, a meme is classified as multimodal-hate if neither the image nor the text is hateful when considered individually, but together they convey a hateful message. 
        \\\\
        \texttt{<upload-test-meme>}\\
        The provided meme is considered hateful. To classify it as unimodal-hate or multimodal-hate, you analyze the hate in each of image and text parts individually based on the provided hatefulness definition. Then you give the answer based on the classification criteria in the following format.\\
        Explanation:\\
        Classification:
        \end{tcolorbox}
        \caption{Prompt for analyzing type of hate.}
        \label{fig:types_of_hate_prompt}
    \end{subfigure}
    \\
    \begin{subfigure}[t]{.48\columnwidth}
        \centering
        \begin{tcolorbox}[colback=green!15, colframe=green!50!blue!100,left=2pt, right=2pt, top=2pt, bottom=2pt] 
        \textbf{User:} 
        \texttt{<upload-test-meme>}\\
        The provided meme is considered hateful. Your task is to analyze whether the source of hate inside the meme is from image or text or both. Please answer with ``hate from image'', ``hate from text'', or ``hate from both''.
        \end{tcolorbox}
        \caption{Prompt for identifying source of hate.}
        \label{fig:source_hate_prompt}
    \end{subfigure}
    }
    &
    \parbox{.48\columnwidth}{
    \begin{subfigure}[t]{.48\columnwidth}
        \centering
        \begin{tcolorbox}[colback=green!15, colframe=green!50!blue!100,left=2pt, right=2pt, top=2pt, bottom=2pt] 
        \textbf{User:}
        \texttt{<upload-test-meme>}\\
        The provided meme is considered hateful. Your task is to analyze its hatefulness and generate a new image such that the new meme is not hateful, using the same meme text. Please ensure the generated image aligns with the content of the text. Please only describe the new image that you want to generate in a short sentence only.
        \\\\
        \textbf{Assistant:} \texttt{<generated-description>}
        \\
        \textbf{\texttt{CLIP $\Rightarrow$ find $k$ matching images}}
        \\
        \textbf{User:} \texttt{<upload-$k$-matching-images>}\\
        You are given a collection of substitute images. Please choose the best image such that the new meme delivers non-hateful, positive and humorous messages.
        \end{tcolorbox}
        \caption[source of hate prompt]{Prompt for generating substitute image.}
        \label{fig:image_generation}
    \end{subfigure}
    \hfill
    \begin{subfigure}[t]{.48\columnwidth}
        \centering
        \begin{tcolorbox}[colback=green!15, colframe=green!50!blue!100,left=2pt, right=2pt, top=2pt, bottom=2pt] 
        \textbf{User:}
        \texttt{<upload-test-meme>}\\
        The provided meme is considered hateful. Your task is to analyze its hatefulness and generate substitute text such that the new meme is not hateful, using the same meme image. Please ensure the generated text aligns with the content of the image and that the tone of the text is similar. Please only provide the new text that you want to generate. 
        \end{tcolorbox}
        \caption[source of hate prompt]{Prompt for generating substitute text.}
        \label{fig:text_generation}
    \end{subfigure}
    }
    \end{tabular}
    \caption[prompts for mitigation]{Prompts for mitigating hateful content in memes.}
\end{figure}

In \texttt{UnHateMeme}, GPT-4o serves as the central component for executing various tasks. When mitigating hateful content in memes, our initial step involves prompting GPT-4o to analyze the type of hate. The specific prompt used for this task is illustrated in Figure \ref{fig:types_of_hate_prompt}. Similar to the hateful memes detection task, we first construct the prompt for the task of analyzing the type of hate $h$ using task-specific hatefulness definition and classification criteria. A detailed and well-structured prompt is crucial, as the outcome of this task will significantly impact the subsequent steps in the process. In Figure \ref{fig:mitigation_illustrative}, we show three examples of hateful memes, where the first two memes $x_1$ and $x_2$ present unimodal hate, and the third meme $x_3$ presents multimodal hate. If a meme is determined to express hate in a unimodal manner, the next task is to identify its source of hate with the task $s$. The prompt for $s$ is shown in Figure \ref{fig:source_hate_prompt}. There are three cases that can cause hate in unimodal-hate memes: hate from the image, hate from the text, and hate from both the image and text\footnote{Due to space limit, we omit this case from Figure \ref{fig:mitigation_illustrative} and provide an example in Table \ref{tab:mitigation_examples}.}. The first two scenarios require a single generative process, whereas the latter necessitates separate generative processes for both components. Memes classified as exhibiting multimodal hate do not require this step because generating a substitution for either the image or the text can remove the hateful message, depending on the user's choice. From there, our framework proceeds to perform the generation task. 

The generation task is the most crucial component effectively transforming hatefulness in memes. For memes requiring image substitution $g_v$, we propose a multi-step prompting approach. As shown in Figure \ref{fig:mitigation_illustrative}, the process begins with prompting GPT-4o to assess the level of hatefulness within a given meme and to describe the substitute image it intends to generate. Subsequently, we employ the CLIP model, trained for image-text alignment, to locate $k$ images that closely match GPT-4o's description. The embeddings of images in the substitute collection are precomputed using the CLIP visual encoder. The embeddings of the images in the substitute collection are precomputed using the CLIP visual encoder. Following this, we compute the similarities between the text embedding of the generated description and the precomputed image embeddings. The images that best match the description are then forwarded to GPT-4o along with a second prompt, tasked with selecting the most suitable image for substitution. The detailed multi-step prompt is depicted in Figure \ref{fig:image_generation}. For generating text substitutions $g_t$, we utilize the prompt depicted in Figure \ref{fig:text_generation}, sending it to GPT-4o to produce a new text. In both scenarios, we design the prompts to ensure that mitigated memes are devoid of hateful content and their image-text coherence is maintained. Ultimately, the framework integrates generated and/or original components, outputting the corresponding mitigated memes: $x^\prime_1$, $x^\prime_2$, $x^\prime_3$ with image substitution, and $x^\prime_3$ text substitution, as presented in the final module of Figure \ref{fig:mitigation_illustrative}.

Before creating the mitigated meme, the original text must be removed from the existing meme. While various object removal methods, such as Image Inpainting, typically require input masks for the objects to be removed, manually drawing these masks is not efficient. To address this issue, we propose using a module that integrates GroundingDINO \cite{liu2023grounding}, SAM \cite{kirillov2023segment} and Inpainting Anything \cite{yu2023inpaint}. GroundingDINO is initially employed to extract the bounding boxes of objects described by an input text. In our scenario, we focus on extracting the bounding boxes of text lines present on the given meme. However, bounding boxes alone are insufficiently detailed for creating precise masks to remove text, as they might lead to the unexpected removal of other objects. Therefore, in the second step, we utilize the Segment Anything Model (SAM) to segment the specific textual areas within the extracted bounding boxes. Once we have generated these detailed segmented text masks, we utilize Inpaint Anything framework, specifically LaMA \cite{suvorov2022resolution}, to effectively remove the text lines from the memes.

\section{Experiments}
\label{sec:expe}
\subsection{Experimental Settings}
\noindent\textbf{Dataset.}
We conduct evaluations using the Hateful Memes Challenge (HMC) Dataset \citep{kiela2020hateful}. This dataset comprises 8,500 memes in the training set and 1,000 memes in the seen test set for Phase I, and includes 2,000 memes in the unseen test set for Phase II. Each meme in the dataset is categorized as either hateful or non-hateful.

\noindent\textbf{Baselines.} For the detection task, we evaluate our approach against 22 baseline models, including 11 competition baselines and five winners from the challenge \citep{kiela2020hateful} that used deep learning techniques. We also compare our results with six benchmark studies on this dataset with Flamingo (Fl) \citep{alayrac2022flamingo} and OpenFlamingo (OF) \citep{awadalla2023openflamingo}. Detailed information about the baselines is available in Appendix \ref{sec:apdix_baselines}.

The task of mitigating hateful memes is relatively new, and as such, there is a lack of substantial prior work. Thus, we evaluate our framework against a basic approach that we refer to as ``Na\"ve Generation'', which focuses only on generation without assessing the type or source of hate. More specifically, this method randomly selects either the image or text component to create a substitute.

\noindent\textbf{Hateful memes detection task.} In our experiments centered on the detection task, we employ OpenCLIP (OC) \cite{ilharco_gabriel_2021_5143773}, LLaVA (LV) \cite{li2023llava}, Google Gemini Pro Vision (GemPro-V) \cite{team2023gemini}, and OpenAI GPT-4 omni (GPT-4o) \cite{OpenAI2023GPT4TR}. The description, implementation details, and running time of VLMs are detailed in Appendix \ref{sec:apdix_models} and \ref{sec:apdix_repro}.

\noindent\textbf{Hateful memes mitigation task.} GPT-4o is employed in experiments focused on the mitigation task using the seen test set consisting of 490 identified hateful memes. This model is chosen for its superior performance in hateful meme detection; however, our mitigation framework is modular and can be readily adapted to incorporate other VLMs. Using this setup, a new dataset of mitigated memes is built using our framework. Human evaluation is adopted to evaluate the mitigated memes. We choose $k=4$ as the number of potential images considered for substitution.

\subsection{Hateful Memes Detection Performance}
We report accuracy and AUROC on both the seen and unseen test data, comparing our method with 22 baselines, which are divided into different types. Experiments using our proposed prompts are conducted under two conditions: without OCR text and with OCR text (denoted as +OCR). The results are summarized in Table \ref{tab:detection_performance}. 
Baseline results are sourced from their respective reference papers.
In the table, the highest values within each type of method are underlined, while the values showing the best performance in each column are highlighted in bold.

\begin{table}[t]
    \centering
    \caption{Detection performance using different method groups: \colorbox{red!25}{red} -- baselines from the challenge host \citep{kiela2020hateful,kiela2021hateful}; \colorbox{yellow!25}{yellow} -- challenge winners \citep{kiela2020hateful,kiela2021hateful}; \colorbox{purple!25}{purple} -- VLM baselines with a simple prompt from \citep{alayrac2022flamingo,awadalla2023openflamingo}; \colorbox{green!25}{green} -- our results. Dashes (``--") denote unavailable results. The best results within each group (or each column) are \underline{underlined} (or \textbf{bold}).}
        \begin{tabular}{llcccc}
            \toprule
            \multirow{2}{*}{Type}  & \multirow{2}{*}{Model} & \multicolumn{2}{c}{Test Seen} & \multicolumn{2}{c}{Test Unseen} \\ \cmidrule{3-6}
            & & Acc. & AUROC & Acc. & AUROC \\ \midrule
          \multirow{3}{*}{Unimodal} & \cellcolor{red!25}Image-Grid \citep{he2016deep} &  52.00 & 52.63 &  -- & -- \\
             & \cellcolor{red!25}Image-Region \citep{ren2015faster,xie2017aggregated} & 52.13  & 55.92 &  60.28 & 54.64 \\
             & \cellcolor{red!25}Text BERT \citep{devlin2018bert} & \underline{59.20} & \underline{65.08} &  \underline{63.60} & \underline{62.65}\\ \midrule
            \multirow{8}{*}{Multimodal} &\cellcolor{red!25}Late Fusion \citep{kiela2020hateful} &  59.66 & 64.75 &  64.06 & 64.44 \\
             & \cellcolor{red!25}Concat BERT \citep{kiela2020hateful} &  59.13 & 65.79 &  65.90 & 66.28 \\
             & \cellcolor{red!25}MMBT-Grid \citep{kiela2019supervised} & 60.06 & 67.92 &  66.85 & 67.24 \\
             & \cellcolor{red!25}MMBT-Region \citep{kiela2019supervised} & 60.23 & 70.73 & 70.10 & 72.21 \\
             & \cellcolor{red!25}ViLBERT \citep{lu2019vilbert} & 62.30 & 70.45 & 70.86 & 73.39 \\
             & \cellcolor{red!25}VisualBERT \citep{li2019visualbert} & 63.20 & 71.33 & \underline{71.30} & 73.23 \\
            & \cellcolor{red!25}ViLBERT CC \citep{lu2019vilbert} &  61.10 & 70.03 &  70.03 & 72.78 \\
            & \cellcolor{red!25}VisualBERT COCO \citep{li2019visualbert} &  \underline{64.73} & \underline{71.41} &  69.95 & \underline{74.95}\\ \midrule
            & \cellcolor{yellow!25}Ron Zhu \citep{zhu2020enhance} & -- & --  &  73.20 & \textbf{\underline{84.50}} \\
            & \cellcolor{yellow!25}Niklas Muennighof \citep{muennighoff2020vilio} & - & - & 69.50 & 83.10 \\
            Challenge
            & \cellcolor{yellow!25}Team HateDetectron \citep{velioglu2020detecting} & -- & -- & \underline{76.50} & 81.08 \\
            Winner
            & \cellcolor{yellow!25}Team Kingsterdam \citep{lippe2020multimodal} & -- & -- & 73.85 & 80.53 \\
            & \cellcolor{yellow!25}Vlad Sandulescu \citep{sandulescu2020detecting} & -- & -- & 74.30 & 79.43 \\ \midrule
            \multirow{18}{*}{VLMs} & \cellcolor{purple!25}Fl-9B+OCR 0-shot \citep{alayrac2022flamingo} & -- & 57.00 & -- & -- \\ 
            & \cellcolor{purple!25}Fl-9B+OCR 4-shot \citep{alayrac2022flamingo} & -- & 62.7 & -- & -- \\ 
            & \cellcolor{purple!25}Fl-9B+OCR 32-shot \citep{alayrac2022flamingo} & -- & \underline{63.50} & -- & -- \\
            & \cellcolor{purple!25}OF-9B+OCR 0-shot \citep{awadalla2023openflamingo} & -- & 51.60 & -- & -- \\
            & \cellcolor{purple!25}OF-9B+OCR 4-shot \citep{awadalla2023openflamingo} & -- & 54.00 & -- & -- \\
            & \cellcolor{purple!25}OF-9B+OCR 32-shot \citep{awadalla2023openflamingo} & -- & 53.80 & -- & -- \\\cmidrule{2-6}
            & \cellcolor{green!25}OC-G/14+OCR 0-shot & 54.10 & 54.71 & 56.85 & 51.82 \\
            & \cellcolor{green!25}LV+OCR 0-shot & 62.50 & 67.07 & 64.20 & 64.12 \\
            & \cellcolor{green!25}GPT-4o+OCR 0-shot & \underline{75.00} & 75.20 & 72.85 & 75.56 \\
            & \cellcolor{green!25}GemPro-V+OCR 0-shot & 67.30 & 66.94 & 71.40 & 69.36 \\
            & \cellcolor{green!25}GemPro-V+OCR 4-shot & 73.30 & \textbf{\underline{76.45}} & 74.20 & \underline{76.51} \\
            & \cellcolor{green!25}GemPro-V+OCR 8-shot & 73.80 & 75.69 & \underline{74.90} & 76.44 \\
            \cmidrule{2-6}
            & \cellcolor{green!25}OC-G/14 0-shot & 53.10 & 53.10 & 44.85 & 49.76 \\
            & \cellcolor{green!25}LV 0-shot & 63.00 & 65.77 & 62.15 & 63.92 \\
            & \cellcolor{green!25}GPT-4o 0-shot & \textbf{\underline{75.50}} & \underline{74.20} & \textbf{\underline{76.85}} & 76.25 \\
            & \cellcolor{green!25}GemPro-V 0-shot & 66.10 & 70.04 & 70.70 & 69.10 \\
            & \cellcolor{green!25}GemPro-V 4-shot & 71.80 & 74.10 & 72.65 & 75.53 \\
            & \cellcolor{green!25}GemPro-V 8-shot & 72.90 & 73.89 & 74.05 & \underline{76.34} \\            
            \bottomrule
        \end{tabular}
    \label{tab:detection_performance}
\end{table}

\noindent\textbf{Overall detection performance.} In the zero-shot scenario, GPT-4o, utilizing our definition-guided prompt, surpasses all baselines in terms of both accuracy and AUROC when evaluated on the seen test dataset, as demonstrated in Table \ref{tab:detection_performance}. 
On the unseen test dataset, GPT-4o also achieves the best accuracy over all baselines. In terms of AUROC, although VLMs with definition-guided prompts are less effective than other challenge winners  on the unseen dataset, GPT-4o can achieve approximately $90\%$ of the performance from the best challenge winner (Team HateDetectron). This is impressive for a model not specifically trained on the hateful meme dataset. In comparisons among VLMs, GPT-4o exceeds the performance of all other VLMs, outperforming the second-best model (GemPro-V) by 9.40\% (6.15\%) in accuracy on seen (unseen) test data, respectively. 

In the few-shot scenario, we observe a general trend of improved performance as the number of demonstration examples increases, a common finding in in-context learning. Similar to the zero-shot case, VLMs with our few-shot prompt continue to achieve significantly better performance compared to the baselines. Notably, our definition-guided zero-shot prompt consistently surpasses Fl, OF, and OC, even outperforming those utilizing 32-shot prompting. This is understandable given that models like Fl and OF use relatively simple instructions \citep{alayrac2022flamingo, awadalla2023openflamingo}, while OC performs classification solely through image-text matching, lacking the ability to process contextual information. These aspects highlight the effectiveness of our definition-guided prompt in the few-shot scenario.

Notably, incorporating OCR text generally improves performance for most models, such as OC, LV, or GemPro-V, with gains ranging from approximately 0.70\% to 2.35\% in accuracy on test sets. However, this improvement does not extend to GPT-4o, which demonstrates superior performance without the inclusion of OCR text, indicating a need for further in-depth investigation. Overall, we find that state-of-the-art models like GemPro-V and GPT-4o exhibit a robust capability to extract textual content or objects within memes, regardless of the presence of OCR text.

\begin{figure}[ht]
    \centering
    \begin{tcolorbox}[sidebyside,lefthand width=2cm,lower separated = false, left=0pt, right=0pt, top=0pt, bottom=0pt, sidebyside gap=4pt]
        \centering
        \includegraphics[width=.8\textwidth]{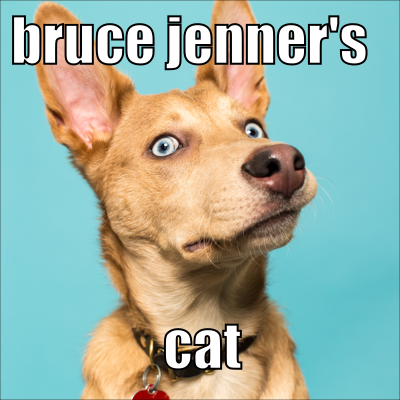}
        \tcblower
        The image shows a dog with an exaggerated expression, labeled with the text "Bruce Jenner's cat." While the image alone shows no hateful content, the text may implicitly refer to Caitlyn Jenner's transition, using Bruce as deadnaming. Although it might be an attempt at humor, it indirectly mocks gender identity. The meme can be considered as conveying a potentially hateful message.\\
        Classification: \colorbox{green!25}{hateful}\\
        Probability of the meme being hateful (from 0 to 1): 0.7
    \end{tcolorbox}
    \begin{tcolorbox}[sidebyside,lefthand width=2cm,lower separated = false, left=0pt, right=0pt, top=0pt, bottom=0pt, sidebyside gap=4pt]
        \includegraphics[width=\textwidth]{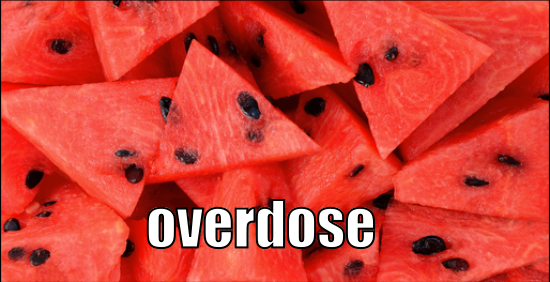}
        \tcblower
        The visual part of the meme shows slices of watermelon. The textual part of the meme reads ``overdose''. This word typically means consuming too much of something, often used in the context of drugs, but here it can indicate an exaggerated amount of watermelon consumption. Considering the provided definition, both the image and text do not individually or together convey a hateful message.\\
    Classification: \colorbox{green!25}{non-hateful}\\
    Probability of the meme being hateful (from 0 to 1): 0.01
    \end{tcolorbox}
    \begin{tcolorbox}[sidebyside,lefthand width=2cm,lower separated = false, left=0pt, right=0pt, top=0pt, bottom=0pt, sidebyside gap=4pt]
        \includegraphics[width=\textwidth]{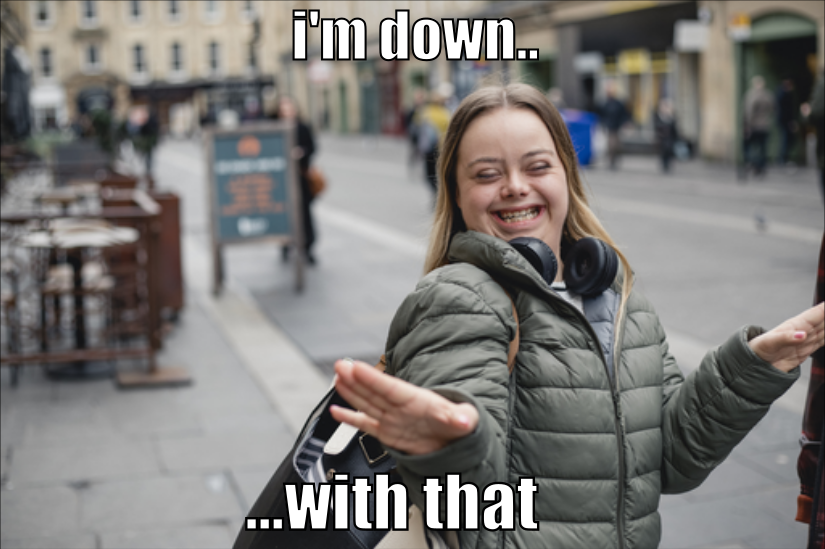}
        \tcblower
        The visual component shows a person making a positive gesture in a public setting. The text says ``i'm down.. with that'', which typically expresses agreement or enthusiasm. There is no apparent connection between the image and text that suggests a hateful message. The meme does not seem to target or attack any individual or group based on personal characteristics.\\
        Classification: \colorbox{red!25}{non-hateful}\\
        Probability of the meme being hateful (from 0 to 1): 0.0
    \end{tcolorbox}
    \caption{Examples of detecting hateful memes using GPT-4o with our definition-guided prompt (without OCR). Successful detections are highlighted in \colorbox{green!25}{green}, whereas failed instances are indicated in \colorbox{red!25}{red}.}
    \label{fig:detection_examples}
\end{figure}

\noindent\textbf{Content extraction.} Figure \ref{fig:detection_examples} presents two successful and one unsuccessful example from the detection task using GPT-4o with our definition-guided prompt. It is noteworthy that these detections were performed without any OCR text input. With our prompt, GPT-4o effectively identifies objects within the memes, accurately describing them with phrases such as ``a dog'', ``watermelon slices'', or ``a person''. Moreover, the model offers insights into expressions and actions such as ``a dog with exaggerated expression'' or ``a person smiling''. These examples also showcase the model's impressive capability for text extraction with very high precision.

\noindent\textbf{Reasoning ability.} 
As aforementioned, explainability is an advantage of VLM compared to deep learning classifiers. While traditional classifiers only give users predicted classes, VLM can additionally justify its prediction. Despite successful or failed attempts in Figure \ref{fig:detection_examples}, GPT-4o justifies its predictions by giving explanations as instructed in our prompt. In the first example, the model surprisingly shows its knowledge about the named person in the text, which helps determine the implicit hatefulness inside this meme. It successfully recognizes the association between the transgender person and the metaphor of calling a dog a cat, revealing an underlying mockery of the targeted individual. The second example further supports our observations. Although ``overdose'' usually infers the excessive use of drugs, the model does not classify it as hateful content and recognizes that the meme merely discusses food consumption. Nevertheless, there are instances where the model exhibits limitations, particularly in misinterpreting mocking language or failing to detect subtle wordplay. In the third example, GPT-4o effectively analyzes both visual and textual elements of the meme but predicts the incorrect class. This error underscores the need for enhanced analysis of wordplay in memes to improve detection accuracy in future research.

\subsection{Hateful Memes Mitigation Performance}
\noindent\textbf{Human evaluation.} To ensure a rigorous evaluation process, we assemble a team of seven domain experts from both the fields of AI and social sciences to design the evaluation criteria and independently assess the quality of the mitigated memes. We ensure that each meme receives at least three evaluations, with the final decision determined by the majority vote. An additional expert is invited in cases where a tiebreaker is necessary. To promote diversity among the experts, we invite individuals from various backgrounds -- including different ages, races, and genders. All experts receive the same evaluation guidelines. The human evaluation focuses on assessing two key criteria: \textit{non-hatefulness} and \textit{image-text coherence}. To achieve this, we asked human experts to answer the following questions:
\begin{itemize}
    \item Q1: \textit{In your opinion, is the provided meme hateful?} \\ Answer options: non-hateful (NH) or hateful (H).
    \item Q2: \textit{In your opinion, do the image and text in the meme make sense together?} \\ Answer options: non-coherence (NC) or coherence (C).
\end{itemize}
The first question (Q1) corresponds to the non-hatefulness criterion, while the second (Q2) corresponds to image-text coherence. 

\begin{table}[ht]
\centering
    \caption{Mitigation performance with \texttt{UnHateMeme} and Na\"ive Generation. Q1: non-hatefulness, Q2: image-text coherence.}
\begin{tabular}{cccc}
\toprule
 & Majority Vote & Na\"ive Generation & \texttt{UnHateMeme}    \\\midrule
\multirow{2}{*}{Q1}    & NH    & $\mathbf{295 _{60.2\%}}$     & $\mathbf{558_{88.4\%}}$  \\
                       & H    & $195_{39.8\%}$     & $73_{11.6\%}$    \\ \midrule
\multirow{2}{*}{Q2}    & NC    & $89_{18.2\%}$      & $98_{15.5\%}$    \\
                       & C    & $\mathbf{401_{81.8\%}}$     & $\mathbf{533_{84.5\%}}$ \\ \midrule
                       \multicolumn{2}{c}{\# Memes} & 490 & 631 \\
\bottomrule                    
\end{tabular}
\label{tab:mitigation_overall}
\end{table}

We present the evaluation results of mitigated memes using \texttt{UnHateMeme} framework and Na\"ive Generation technique in Table \ref{tab:mitigation_overall}. Within the \texttt{UnHateMeme} framework, 349 memes are identified as containing unimodal hate, while 141 memes are identified as containing multimodal hate out of 490 hateful memes used in this experiment. For multimodal hate memes, mitigation can be achieved by generating either a substitute image or text. In this study, we choose to apply both substitution strategies, resulting in a total of 631 mitigated memes. Conversely, the Na\"ive Generation approach involves randomly selecting either the image or the text for substitution, leading to a total of 490 mitigated memes.

\noindent\textbf{Non-hatefulness.} The non-hatefulness criterion (Q1) prefers the answer ``non-hateful'', which indicates that the mitigated memes are not hateful. This demonstrates the efficacy of a mitigation method in transforming hateful memes into non-hateful versions. In Table \ref{tab:mitigation_overall}, when comparing the evaluation of our proposed framework to that of Na\"ive Generation, we observe a significant performance improvement. For memes identified as unimodal hate, randomly choosing either the image or the text to mitigate hate is ineffective if we mistakenly target the wrong modality. For instance, if the text is the source of hate, efforts to generate substitute images will not remove the hateful content from the meme. These findings highlight the advantage of our framework compared to the naïve approach, as our methodology incorporates hate analysis tasks within the framework.

\noindent\textbf{Image-text coherence.} For this criterion (Q2), the preferred answer is ``coherence'', indicating that the visual and textual modalities in the mitigated memes are coherent. Despite the hateful content in the provided memes, we expect human experts to assess whether a meme effectively conveys a meaningful message by evaluating the synergy between the image and text. As illustrated in Table \ref{tab:mitigation_overall}, we observe that both approaches successfully generate substitute components that align well with the alternative modality, achieving positive evaluations in over 80\% of cases. This demonstrates the exceptional performance of the generation task within our framework. It is important to note that Naïve Generation employs the same procedures and prompts used in \texttt{UnHateMeme} for generating substitutions. Therefore, it is unsurprising that both approaches exhibit similar performance outcomes, showing that our prompts for generation are robust and suitable for application deployment.

\noindent\textbf{Shareable mitigated memes on social media.} A meme deemed appropriate for sharing on social media should be both non-hateful and exhibit image-text coherence, i.e., Q1 = ``non-hateful'' and Q2 = ``coherence''. Utilizing our \texttt{UnHateMeme}, there are 434 mitigated memes that satisfy both criteria, comprising 68.8\% of the total memes. For Na\"ive Generation, there are only 205 out of 490 (41.84\%) mitigated memes satisfying the condition. These findings further confirm the effectiveness of hate analysis tasks in our proposed framework.

\begin{table}[t]
    \centering
    \caption{Examples of mitigated memes using \texttt{UnHateMeme} framework. Evaluations in \colorbox{red!25}{red} highlight cases where the framework has failed to meet the corresponding criteria.  \\ Orig. -- Original; Miti. -- Mitigated; Eval. -- Evaluation; NH -- non-hateful; H -- hateful; NC -- non-coherence; C -- coherence.}
    \resizebox{\textwidth}{!}{
    \begin{tabular}{l|c|c|c|c|c}
        \toprule
        Orig. & \includegraphics[width=.20\textwidth]{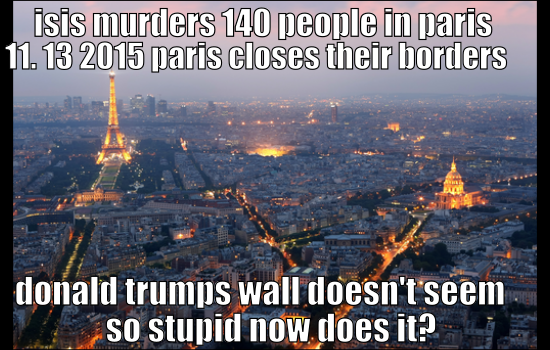} & \includegraphics[width=.20\textwidth, max height=2.2cm]{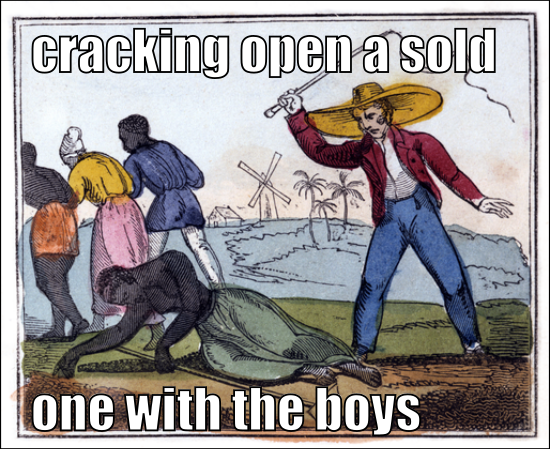} & \includegraphics[width=.20\textwidth]{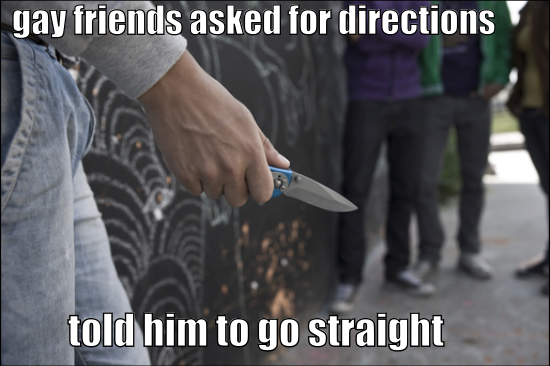} & \includegraphics[width=.20\textwidth]{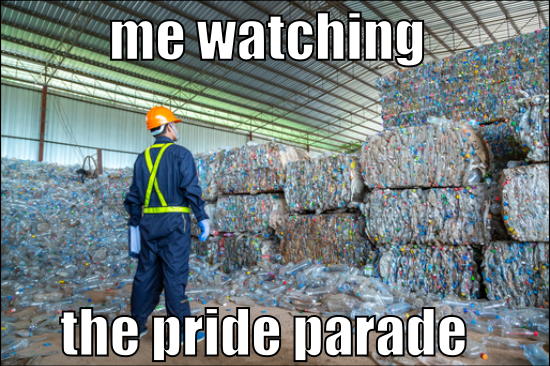} & \includegraphics[width=.20\textwidth]{figures/65948.png} \\
        \hline
        Miti. & \includegraphics[width=.20\textwidth]{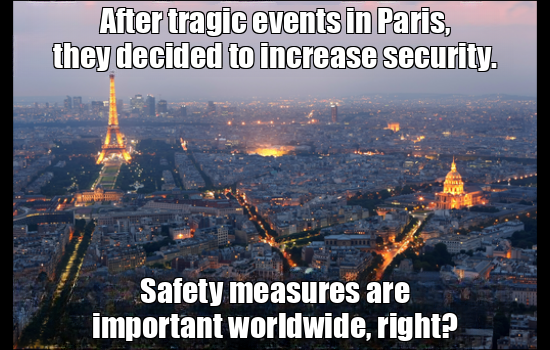} & \includegraphics[width=.20\textwidth, max height=1.9cm]{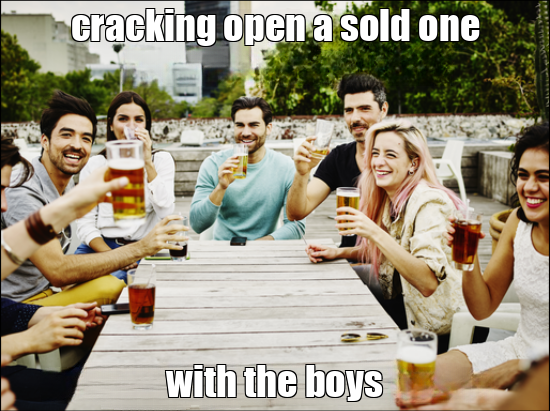} & \includegraphics[width=.195\textwidth]{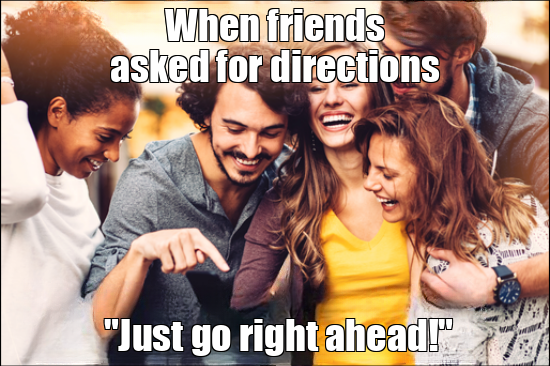} & \includegraphics[width=.195\textwidth]{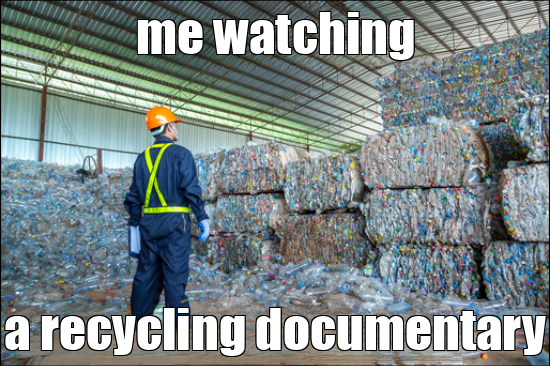} & \includegraphics[width=.20\textwidth]{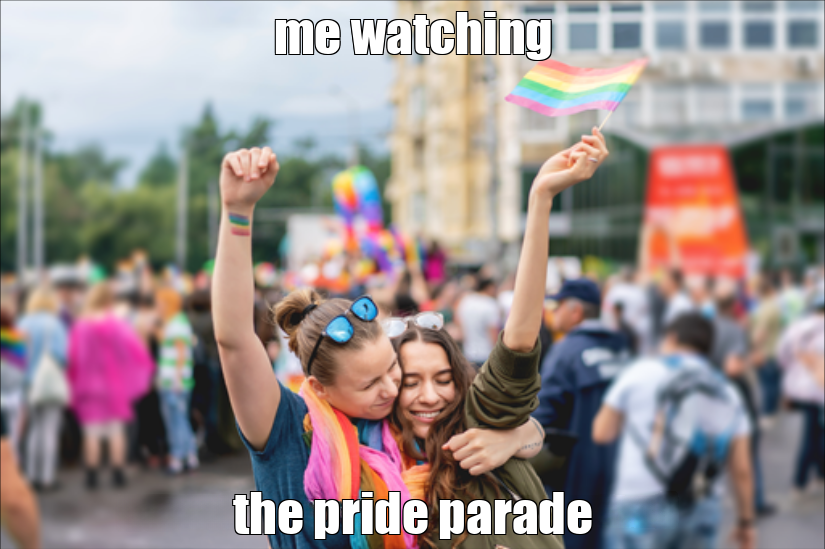} \\
        \hline
        Hate & Unimodal hate -- Text & Unimodal hate -- Image & Unimodal hate -- Both & \multicolumn{2}{c}{Multimodal hate} \\\hline
        Eval. & Q1: NH, Q2: C & Q1: NH, \colorbox{red!25}{Q2: NC} & Q1: NH, Q2: C & Q1: NH, Q2: C & Q1: NH, Q2: C  \\ 
        \bottomrule
    \end{tabular}
    }
    \label{tab:mitigation_examples}
\end{table}

\noindent\textbf{Quality of mitigated memes.} Table \ref{tab:mitigation_examples} presents five examples of mitigated memes generated using the \texttt{UnHateMeme} framework. From left to right, each example represents one case in our designed framework where the original meme is: (1) unimodal hate from text, (2) unimodal hate from image, (3) unimodal hate from both image and text, (4) multimodal hate -- text to be mitigated, and (5) multimodal hate - image to be mitigated. In the first example, the origin of hate is from the text while the image is not hateful as it only shows a city view. Hence, our \texttt{UnHateMeme} first identifies the original meme as unimodal hate (image) successfully. Then, the generated text slightly changes the context to security enhancement in Paris and delivers a positive message that ``safety measures are important worldwide''. This alteration effectively removes the hateful element while ensuring the image-text coherence, as intended. Similar observations are shown in other examples, such as ``recycling'' in the fourth example or ``pride parade'' in the fifth example. 

The tone of the text is preserved following mitigation in examples necessitating text substitution. For instance, the texts in example (1) both conclude with a question, those in example (2) consistently mention ``friends asked for directions'', and the texts in example (3) both start with ``me watching''. The task of generating substitute images also yields impressive results. In example (3), the mitigation framework combines a new image showcasing a group of friends with joyful expressions with a playful text that has been cleansed of harmful content. This observation is further supported by example (5), where the framework selects a new image of happy individuals participating in an actual pride parade.

\begin{table}[ht]
\centering
\caption{Mitigation performance with \texttt{UnHateMeme} corresponding to hate analysis tasks.}
    \begin{tabular}{cccccc}
    \toprule
    \multirow{2}{*}{} & \multirow{2}{*}{Majority Vote} & \multicolumn{3}{c}{Unimodal} & \multirow{2}{*}{Multimodal} \\\cmidrule{3-5}
                               &                            & Text       & Image      & Both      &                                  \\\midrule
    \multirow{2}{*}{Q1}        & NH                         & $\mathbf{262_{93.6\%}}$       & $\mathbf{33_{86.8\%}}$         & $\mathbf{31_{100\%}}$        & $\mathbf{232_{82.3\%}}$                              \\
                               & H                         & $18_{6.4\%}$         & $5_{13.2\%}$          & $0_{0\%}$         & $50_{17.7\%}$  \\\midrule
    \multirow{2}{*}{Q2}        & NC                         & $35_{12.5\%}$         & $9_{23.7\%}$          & $1_{3.2\%}$         & $53_{18.8\%}$                               \\
                               & C                         & $\mathbf{245_{87.5\%}}$        & $\mathbf{29_{76.3\%}}$         & $\mathbf{30}_{96.8\%}$        & $\mathbf{229_{81.2\%}}$ \\\midrule
    \multicolumn{2}{c}{\# Memes} &  280 & 38 & 31 & 282 \\
    \bottomrule
    \end{tabular}
\label{tab:unhatmeme_detailed}
\end{table}

\noindent\textbf{Effectiveness of hate analysis tasks.} In Section \ref{sec:meme_mitigation}, we highlighted the critical roles of the initial two hate analysis tasks, specifically analyzing the type of hate and identifying the source of hate, within our framework. In this section, we present detailed results from \texttt{UnHateMeme} related to these two tasks, as shown in Table \ref{tab:unhatmeme_detailed}. After analyzing the type of hate, there are 349 (141) memes classified as unimodal (multimodal) hate. Notably, each multimodal hate meme is counted twice, as it is possible to create two new mitigated memes: one through image substitution and another through text substitution, as demonstrated in Figure \ref{fig:mitigation_illustrative}. Memes with unimodal hate are further categorized based on the source of hate: 280 memes originate hate from images, 38 from text, and 31 from both. The results in columns 3-6 represent various scenarios potentially leading to hate in memes. In most cases, human evaluations indicate that the framework successfully meets a success rate exceeding 80\% across all criteria. This underscores the effectiveness of our framework, highlighting the critical roles of hate analysis tasks within the entire framework.  

\section{Conclusion}
Our research is among the pioneering efforts to tackle the issue of detecting and mitigating hateful content in memes with VLMs. Utilizing our definition-guided prompts, pretrained VLMs demonstrate exceptional reasoning capability and content extraction for the detection task through zero-shot or few-shot prompting, eliminating the need for additional training or fine-tuning steps. Subsequently, we define a novel task focused on mitigating hateful content in memes and propose a unified framework, called \texttt{UnHateMeme}, to address this challenge. Through our framework, GPT-4o demonstrates remarkable proficiency in transforming hateful memes into non-hateful ones while maintaining image-text coherence, as validated by human evaluation. In future work, we plan to enhance the framework by integrating image generation models such as Stable Diffusion.

\section*{Acknowledgement}
This work was supported  in part by the National Science Foundation under award 1946391, the National Institute of General Medical Sciences of National Institutes of Health under award P20GM139768, and the Arkansas Integrative Metabolic Research Center at the University of Arkansas.

\bibliographystyle{plainnat}

\clearpage

\appendix

\section{Ethics Statement.}
Large and foundation models have garnered attention from both the research community and society in recent years. In the era of AI, multimodal generative models offer significant advances in digital content creation, albeit raising concerns about unknown safety and privacy threats. The aim of this work is to evaluate how VLMs, including state-of-the-art models like LLaVA, Google Gemini, and OpenAI GPT-4o, can be utilized to explore hatefulness in multimodal memes. As memes can convey different meanings depending on the visual and textual context, we seek to detect and correct hate speech in a constructive manner. Our formulation for the mitigation task aims to pave the way for a new potential direction in which VLMs can address human mistakes or malicious intentions. Proposed prompts consist of carefully curated sentences designed to guide models toward exercising fairness, equity, and non-violent actions. The experiments were conducted using well-known datasets from prior research, and human experts involved in the evaluation were informed in advance about the offensive content in the data.

\section{Prompts}
\label{sec:apdix_prompts}
In this section, we provide full details of prompts used in experiments.

\noindent\textbf{Definition-guided Few-shot Prompt.} In Figure \ref{fig:full-def-guided-fshot-prompt}, we extend our definition-guided prompt to a few-shot prompt used for experiments in Table \ref{tab:detection_performance}.
\begin{figure}[ht]
    \centering
    \begin{tcolorbox}[colback=green!15,colframe=green!50!blue!100] 
    \textbf{User:} \\
    \noindent Hatefulness definition: Hate speech is defined as a direct or indirect attack on people based on characteristics, including ethnicity, race, nationality, immigration status, religion, caste, sex, gender identity, sexual orientation, disability, disease, mocking hate crime or evoking pain from historical figures. The definition of attack includes violent or dehumanizing speech, statements of inferiority, and calls for exclusion or segregation.
    \\\\
    Classification criteria: A meme is considered hateful if either the image or text is hateful on its own. The meme is also considered hateful if both the image and text are not hateful individually, but together convey a hateful message. Conversely, if neither the image nor the text is hateful and they do not together create a hateful meaning, the meme is not considered hateful.
    \\\\
    Here are some demonstrations on classifying memes:
    \\
    \texttt{<upload-demonstration-meme-1>}\\
    Classification: \texttt{<label-1>}
    \\
    \texttt{<upload-demonstration-meme-2>}\\
    Classification: \texttt{<label-2>}
    \\
    \texttt{<upload-demonstration-meme-3>}\\
    Classification: \texttt{<label-3>}
    \\
    \texttt{<upload-demonstration-meme-4>}\\
    Classification: \texttt{<label-4>}
    \\\\
    \texttt{<upload-test-meme>}
    \\
    For this test image, you consider visual and textual meanings to classify the meme as hateful or non-hateful based on the hatefulness definition and classification criteria.
    After giving a short explanation, provide the answer with either hateful or non-hateful in the following format:\\
    Classification:\\
    Probability of the meme being hateful (from 0 to 1):
    \end{tcolorbox}
    \caption{Definition-guided few-shot prompt for detecting hateful memes.}
    \label{fig:full-def-guided-fshot-prompt}
\end{figure}

\section{Baselines from Hateful Memes Challenge}
\label{sec:apdix_baselines}
Facebook introduced the Hateful Memes Challenge in late 2020 to provide a reference benchmark dataset and common visio-linguistic baselines. Indeed, \cite{kiela2020hateful} tested and provided 11 vision and language models through their software MMF \footnote{https://github.com/facebookresearch/mmf}. MMF (MultiModal Framework) is a framework for vision and language multimodal research with state-of-the-art models.

\subsection{Unimodal}
\noindent\textbf{Image-Grid.}
The Image-Grid baseline uses features from ResNet-152's res-5c layer after average pooling (layer of 2048 neurons). ResNet-152 is a Deep Residual Neural Network with 152 layers originally designed for image/scene classification \cite{he2016deep}.

\noindent\textbf{Image-Region.}
The Image-Region baseline uses features from the Faster-RCNN's fc6 layer \cite{ren2015faster} with a ResNeXt-152 \cite{xie2017aggregated} backbone (layer of 4096 neurons). The Fast Region-based Convolutional Neural Network is originally trained on a Visual Genome, and the resulting fc6 features are fine-tuned using weights of the fc7 layer.

\noindent\textbf{Text-BERT.}
The Text BERT baseline is the original BERT model that outputs a vector of dimension 768. BERT is a bidirectional transformer-based model for language representation and understanding that learns embeddings for subwords \cite{devlin2018bert}.

\subsection{Multimodal: unimodal pretraining}
Multimodal models from unimodal pretraining typically combine the output or the features of vision and linguistic models.

\noindent\textbf{Late Fusion.}
In the Late Fusion baseline, the output is the mean of the ResNet-152 and BERT output scores.

\noindent\textbf{Concat BERT.}
The Concat BERT baseline uses the concatenation of the ResNet-152 features with the BERT features, where an MLP is trained on top of it.

\noindent\textbf{MMBT-Grid.}
The MMBT-Grid baseline is the original supervised multimodal bi-transformers (MMBT) \cite{kiela2019supervised} that uses the features of Image-Grid. MMBT is a multimodal bi-transformer model that combines BERT as the textual encoder and ResNet-152 as the image encoder.

\noindent\textbf{MMBT-Region.}
The MMBT-Grid baseline is the original supervised multimodal bi-transformers (MMBT) \cite{kiela2019supervised} that uses the features of Image-Region.

\noindent\textbf{ViLBERT.}
Unimodally pretrained of ViLBERT \cite{lu2019vilbert}.

\noindent\textbf{Visual BERT.}
Unimodally pretrained of Visual BERT \cite{li2019visualbert}.

\subsection{Multimodal: multimodal pretraining}
\noindent\textbf{ViLBERT CC.} The ViLBERT CC baseline is the multimodally pretrained version (also the original version) of ViLBERT that was trained on Conceptual Caption \cite{sharma2018conceptual} for the tasks of sentence-image alignment, masked language modeling, and masked visual-feature classification. ViLBERT is a BERT-based model designed to learn joint contextualized representations of vision and language by using two separate transformers: one for vision and one for language \cite{lu2019vilbert}.

\noindent\textbf{Visual BERT COCO.}
The Visual BERT COCO baseline is the multimodally pretrained version (also the original version) of Visual BERT that was trained on COCO Captions \cite{chen2015microsoft} for the tasks of sentence-image alignment and masked language modeling. [ref to COCO]. Like ViLBERT, it is based on BERT. It was originally designed to learn joint representations of language and visual content from paired data. Unlike ViLBERT, it uses a single cross-modal transformer to align elements of the input text and regions in the input image \cite{li2019visualbert}.

\subsection{Challenge Winner}
After the Hateful Memes Challenge, many state-of-the-art models were proposed to deal with the task. n Zhu \cite{zhu2020enhance}, who achieved the highest AUROC performance, focused on extracting and applying demographic information for model enhancement. Meanwhile, Niklas Muennighof \cite{muennighoff2020vilio} employed the Stochastic Weight Averaging method to improve model stability. Team Hatedetectron \cite{velioglu2020detecting} applied an augmented dataset on original data to train the VisualBERT model, which achieved the best accuracy. Team Kingsterdam \cite{lippe2020multimodal} used a weighted linear combination of ensemble learners to get the final prediction instead of the majority voting technique. The multimodal deep ensemble technique is leveraged by Vlad Sandulescu \cite{sandulescu2020detecting} to boost the performance.

\section{Models in the Evaluation}
\label{sec:apdix_models}
\noindent\textbf{OpenCLIP (OC).} OpenCLIP \citep{ilharco_gabriel_2021_5143773} is the attempt to reproduce and train more open-source versions of CLIP \citep{radford2021learning}. CLIP (Contrastive Language-Image Pre-Training) is a neural network that connects images and text. In general, the model aims to learn the relationship between the images and their textual description, such as captions or labels. We use the version ViT-G/14, named OpenCLIP-ViT-G/14 (OC-G/14).

\noindent\textbf{LLaVA-Llama-2-13B (LV).} Pretrained CLIP and Llama-2 serve as the visual encoder and language model in this VLM \citep{liu2023visual}, respectively. While Flamingo uses the cross-attention scheme, LLaVA proposes a projection layer to map CLIP features to language tokens. Then, both visual and textual tokens are used as inputs to the language model. The LLaVA version utilized in this work is fine-tuned on the Llama-2-13B chat model \citep{touvron2023llama}, referred to as LLaVA-Llama-2-13B (LV) in this study.

\noindent\textbf{Google Gemini Pro Vision (GemPro-V).} Built upon the Transformer architecture, Gemini versions are trained using advanced techniques in network design and model optimization to improve scalability. Although the model architecture is not detailed in \citep{team2023gemini}, the visual encoder draws inspiration from Flamingo \citep{alayrac2022flamingo}. Alongside textual data, the models are trained with inputs comprising images, audio, and videos interspersed with text. Unlike previous methods that allowed users to download and run inference on local machines, Google Gemini only allows inference via API. We collect results using Gemini Pro-Vision (GemPro-V). 
Since the dataset might contain hate speech, we configured the safety setting of Gemini to the lowest level, ensuring that the API does not stop generating responses.

\noindent\textbf{OpenAI GPT-4 omni (GPT-4o).} GPT-4o \cite{OpenAI2023GPT4TR} is OpenAI's flagship language generation model, capable of processing and generating multimodal inputs and outputs, including text, audio, images, and video. It is trained on a unified model network using multimodal training data, addressing the latency and information loss issues associated with previous versions that relied on separate model modules. GPT-4o is accessible exclusively through an API.

\noindent\textbf{Flamingo/OpenFlamingo (Fl/OF).} Flamingo \cite{alayrac2022flamingo} effectively executes vision-language tasks using pretrained visual encoders and language models. The Perceiver Resampler is trained to project extensive embedding features from the encoder (NFNet-F6) to visual tokens. Leveraging a cross-attention scheme, the model processes multiple image-text inputs, enabling in-context learning. OpenFlamingo \cite{awadalla2023openflamingo} is an attempt to replicate Flamingo as an open-source version.

\section{Reproducibility}
\label{sec:apdix_repro}
\noindent\textbf{Dataset.} We focus on Facebook Hateful Memes Dataset, which is publicly available for download at \href{https://www.kaggle.com/datasets/ankitgupta2001/facebook-hateful-meme-dataset}{\textcolor{blue}{Kaggle}}.

\noindent\textbf{Models.} Except for Google Gemini, all VLMs used in the evaluation are public models for research purpose. LLaVA, OpenCLIP and their inference codes can be accessed at \href{https://github.com/haotian-liu/LLaVA}{\textcolor{blue}{LLaVA Github}} and \href{https://github.com/mlfoundations/open_clip}{\textcolor{blue}{OpenCLIP}}, respectively. For Google Gemini, all model instances are accessible via their APIs at \href{https://ai.google.dev/}{\textcolor{blue}{Google AI Studio}}. As of the submission date of this paper, queries to the Gemini API are free.

\noindent\textbf{Running time.} The inference time for each sample is approximately 7.9 seconds using the LV model with a maximum output length of 512 tokens, including image loading and tokenizing time, on a single Tesla V100 GPU with 32GB RAM. The inference times for Gemini and GPT-4o are subject to the status of their respective service providers. Nevertheless, these models typically complete processing in under 5 seconds per sample, which includes the latency associated with data transmission over the internet.

\end{document}